\documentclass[10pt,twocolumn]{article}
\usepackage[margin=0.70in]{geometry}
\usepackage[T1]{fontenc}
\usepackage[utf8]{inputenc}
\usepackage{newtxtext,newtxmath}
\usepackage{microtype}
\usepackage{graphicx}
\usepackage{booktabs}
\usepackage{array}
\usepackage{caption}
\usepackage{subcaption}
\usepackage{xcolor}
\usepackage{hyperref}
\usepackage{enumitem}
\usepackage{amsmath}
\usepackage{xurl}
\usepackage{float}
\usepackage{stfloats}
\usepackage{placeins}
\usepackage{balance}
\usepackage{titlesec}
\usepackage{abstract}
\hypersetup{colorlinks=true,citecolor=blue,linkcolor=blue,urlcolor=blue}
\titleformat{\section}{\bfseries\large}{\thesection.}{0.4em}{}
\titleformat{\subsection}{\bfseries\normalsize}{\thesubsection}{0.4em}{}
\setlist[itemize]{leftmargin=1.2em,itemsep=0.12em,topsep=0.15em}

\newcommand{\cro}{\mathrm{CRO}}

\title{Edge-specific signal propagation on mature chromophore-region 3D mechanism graphs for fluorescent protein quantum-yield prediction}
\author{
Yuchen Xiong$^{1}$, Swee Keong Yeap$^{1,\ast}$ and Steven Aw Yoong Kit$^{1,\ast}$\\[2mm]
\normalsize $^{1}$China-ASEAN College of Marine Sciences, Xiamen University Malaysia,\\
\normalsize Sepang 43900, Selangor, Malaysia\\[1mm]
\normalsize $^{\ast}$Correspondence: Swee Keong Yeap, \href{mailto:skyeap@xmu.edu.my}{skyeap@xmu.edu.my};\\
\normalsize Steven Aw Yoong Kit, \href{mailto:yoongkit.aw@xmu.edu.my}{yoongkit.aw@xmu.edu.my}.
}
\date{}

\begin{document}
\maketitle

\begin{abstract}
\noindent\textbf{Motivation:} Fluorescent protein quantum yield (QY) is controlled by the mature chromophore and its three-dimensional microenvironment, not by sequence identity alone. Generic protein language models and coarse emission-band averages can capture some global trends, but they do not explicitly model how local physical signals act on specific chromophore regions.

\noindent\textbf{Results:} We present a chromophore-centred mechanism graph algorithm for QY prediction. The method converts each PDB structure into a typed 3D residue graph, applies lightweight mature-state chromophore registration, partitions the chromophore into phenolate, bridge and imidazolinone regions, and propagates physicochemical residue signals through activated contact channels. The resulting representation contains 121 chromophore-centred enrichment features, from which a 52-feature non-identity pool is retained before band-specific ExtraTrees regression. Because each feature is defined by a channel--signal--region tuple, the interpretation is built into the representation rather than added only post hoc. On a 531-protein benchmark, the method achieved the best random cross-validation performance among the primary model-based baselines ($R=0.772\pm0.008$, MAE $=0.131\pm0.002$), exceeding Band mean ($R=0.632$), ESM-C ($R=0.734$) and SaProt ($R=0.731$). The method also ranked first on the most stringent random bright-screening task (Bright P@5 $=0.704$). Under fixed 5-mer-Jaccard homology-controlled evaluation, its advantage was clearest in the most remote bucket ($<0.50$ maximum 5-mer Jaccard similarity to the fixed training set; $R=0.697$ versus $0.633$, $0.575$ and $0.408$ for Band mean, SaProt and ESM-C, respectively), with the strongest overall bright/dark Top-K screening in that regime. Stable selected features formed interpretable band-specific motifs: GFP-like models emphasised aromatic bridge/imidazolinone and clamp asymmetry terms, Red models emphasised charge and clamp balance, and Far-red models emphasised flexibility-risk and bulky-contact features.

\noindent\textbf{Availability and implementation:} Source code, processed feature tables and evaluation scripts are available from the first author upon reasonable request.

\noindent\textbf{Contact:} \href{mailto:yuchenak05@gmail.com}{yuchenak05@gmail.com}

\end{abstract}

\section{Introduction}
Fluorescent proteins (FPs) are central tools for live-cell imaging, biosensing and protein engineering. Their practical usefulness depends strongly on quantum yield (QY), which measures the efficiency with which absorbed photons are converted into emitted fluorescence. QY is difficult to infer from sequence because the physical determinant is not merely which residues are present, but how the mature chromophore is constrained, polarised and shielded by its local three-dimensional environment.

This creates a mismatch between many predictive baselines and the underlying photophysics. Sequence-similarity methods can perform well when labelled close homologues are available, but they offer little extrapolative or mechanistic value. Generic protein language models (PLMs) encode broad evolutionary and structural information, yet they are not designed around the mature chromophore. Band-level averages capture coarse emission-class differences but ignore within-band microenvironmental variation.  A related lesson has emerged from graph-based AI for drug discovery: molecular function is often better predicted when the biochemical environment and interaction context are represented explicitly. Environment-aware graph models for protein--ligand affinity, such as EM-PLA, improve binding-affinity prediction by incorporating biochemical environmental information and non-covalent interaction structure \cite{Xie2025EMPLA}. Interpretable network-medicine models, such as iDPath and GraphSynergy, further show that graph models can be designed not only to predict outcomes but also to expose mechanism paths, pivotal proteins or contribution patterns \cite{Yang2022iDPath,Yang2021GraphSynergy}. These ideas motivate an analogous question in fluorescent-protein engineering: can QY be predicted from a local molecular environment graph around the mature chromophore, while retaining a mechanism-level explanation of which chromophore-region interactions drive brightness or darkness?

Here we introduce a structure-derived, environment-aware mechanism graph algorithm built around four design choices. First, the model moves from sequence identity to a typed 3D residue graph. Second, it performs mature-state chromophore registration so downstream features are anchored to the chemically relevant fluorescent core. Third, it partitions the mature chromophore into phenolate, bridge and imidazolinone regions. Fourth, it propagates physicochemical residue signals through robust contact channels to these regions. These steps produce features that are not generic embeddings; they are explicit measurements of how steric locking, hydrophobic packing, charge-related, hydrogen-bond-capable, aromatic and flexibility-related signals are enriched around functional chromophore regions. In this sense, the method treats brightness prediction as local molecular-environment reasoning around a functional chemical centre rather than as direct sequence-to-property regression.

We evaluate the method against three primary model-based baselines: Band mean, ESM-C and SaProt. The method improves random-CV regression, remains competitive across intermediate homology regimes and becomes strongest in the most difficult remote-homology evaluation. More importantly, its selected features reveal coherent chromophore-region mechanisms that are inaccessible to the baseline models and align with known photophysical themes such as chromophore rigidification, electrostatic tuning and torsional-relaxation suppression. The resulting interpretation is intrinsic to the representation: selected features correspond to concrete channel--signal--region paths, rather than to post-hoc explanations attached after prediction.

\section{Materials and methods}
\subsection{Dataset and evaluation design}
The benchmark contained 531 fluorescent proteins with measured QY and emission maxima curated from FPbase. Input structures were taken from the Protein Data Bank when available and otherwise generated using OpenFold3-predicted models \cite{Lambert2019,Berman2000,OpenFold3Team2025}. Proteins were assigned to three predefined emission bands: GFP-like ($500\leq \mathrm{em}<560$ nm), Red ($580\leq \mathrm{em}<610$ nm) and Far-red ($\mathrm{em}\geq610$ nm). Proteins outside these predefined band intervals were not used for band-specific modelling. These bands were used for stratified modelling because different chromophore classes can have distinct QY determinants.

Two evaluation protocols were used. Random evaluation used five seeds and five folds per seed, with out-of-fold predictions pooled for regression metrics. Feature selection, band-specific model fitting and bright/dark Top-K thresholds were recomputed inside each training fold only.

Homology-controlled evaluation used a fixed train--test split based on 5-mer Jaccard similarity rather than global sequence-alignment identity. For a protein sequence $x$, let
\begin{equation}
\mathcal{K}_5(x)=\{x_t x_{t+1}x_{t+2}x_{t+3}x_{t+4}:t=1,\ldots,L_x-4\}
\end{equation}
denote the set of unique contiguous 5-mers in that sequence. The pairwise 5-mer Jaccard similarity between proteins $i$ and $j$ is
\begin{equation}
J_5(i,j)=
\frac{|\mathcal{K}_5(x_i)\cap \mathcal{K}_5(x_j)|}
{|\mathcal{K}_5(x_i)\cup \mathcal{K}_5(x_j)|}.
\end{equation}
A protein was assigned to the fixed training set if it had at least one close 5-mer neighbour in the complete benchmark:
\begin{equation}
i\in\mathcal{D}_{\mathrm{train}}
\quad\Longleftrightarrow\quad
\max_{j\neq i}J_5(i,j)\geq0.85.
\end{equation}
All remaining proteins were assigned to the fixed homology-control test set,
\begin{equation}
\mathcal{D}_{\mathrm{test}}
=
\{i:\max_{j\neq i}J_5(i,j)<0.85\}.
\end{equation}
For each test protein $k$, its similarity to the fixed training set was then defined as
\begin{equation}
m_k=\max_{j\in\mathcal{D}_{\mathrm{train}}}J_5(k,j).
\end{equation}
The test proteins were grouped into three increasingly difficult buckets:
\begin{equation}
\mathrm{bucket}(k)=
\begin{cases}
0.70\text{--}0.85, & 0.70\leq m_k<0.85,\\
0.50\text{--}0.70, & 0.50\leq m_k<0.70,\\
<0.50, & m_k<0.50.
\end{cases}
\end{equation}
The 0.70--0.85 bucket represents relatively close family-level transfer, the 0.50--0.70 bucket represents more distant superfamily-level transfer, and the $<0.50$ bucket represents the most remote extrapolation setting. The split was fixed before model fitting and used only sequence strings, not QY labels or model predictions. Across random seeds, the train--test membership and bucket assignment were unchanged; the seeds only affected stochastic model fitting, such as ExtraTrees initialisation.

We used 5-mer Jaccard similarity rather than global alignment identity because fluorescent proteins in the benchmark vary in length and may contain insertions, deletions or terminal differences. A local k-mer set similarity is less sensitive to gap placement and length-dependent alignment penalties, while still measuring shared sequence-motif content relevant to family-level relatedness.

Bright and dark Top-K metrics used training-fold thresholds in random CV and fixed-training-set thresholds in homology-controlled evaluation, so held-out proteins did not influence the definition of bright or dark candidates.

The primary baselines were Band mean, ESM-C and SaProt \cite{EvolutionaryScale2024,Su2024}. A nearest-neighbour retrieval baseline was not used as a primary comparator because it is a database lookup strategy rather than a transferable model of QY determinants, and it can obscure the intended comparison between generic representations and chromophore-centred mechanism features.

\subsection{Algorithm overview}
\begin{figure*}[t]
    \centering
    \includegraphics[width=0.98\textwidth]{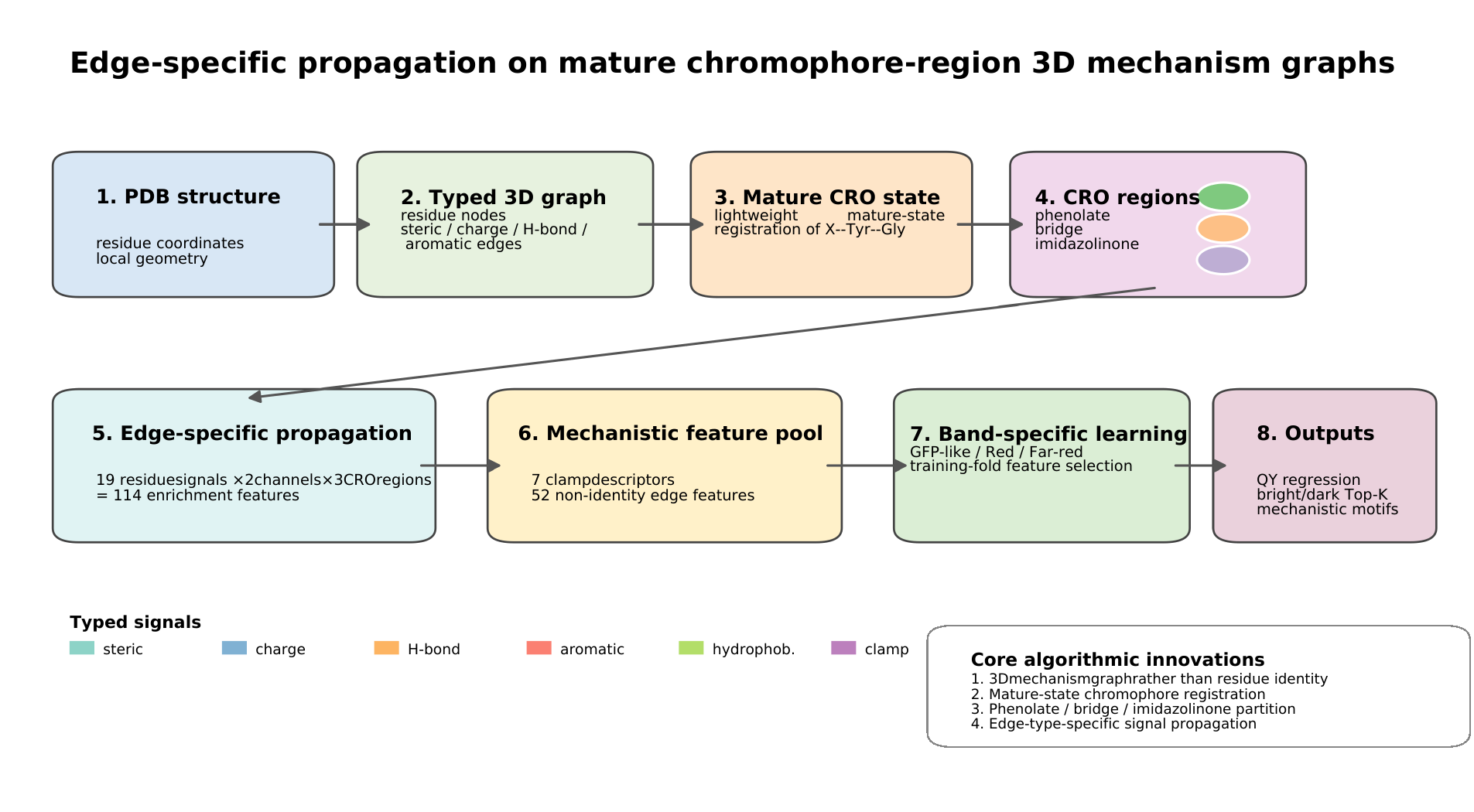}
    \caption{Clean overview of the proposed algorithm. A PDB structure is converted into a typed 3D residue graph, registered to a mature chromophore state, partitioned into functional CRO regions, transformed by edge-specific signal propagation, filtered to remove identity shortcuts and routed to a band-specific predictor.}
    \label{fig:pipeline}
\end{figure*}

The complete algorithm maps a protein structure and emission maximum to a QY prediction through structure parsing, typed graph construction, mature chromophore registration, regional decomposition, edge-specific signal propagation, non-identity feature filtering and band-specific regression (Fig.~\ref{fig:pipeline}). In this benchmark, the structural inputs include OpenFold3-predicted models, so the activated propagation channels are restricted to geometry-robust steric and hydrophobic contacts. The maturation component is intentionally lightweight: it standardises the feature anchor to the mature CRO state and supports region definition, but it is not intended as a quantum-chemical simulation of chromophore formation.

\subsection{Algorithmic novelty}
The proposed method is not a generic residue-level featurisation followed by regression. Its novelty lies in making the functional chemical centre of the protein explicit. First, all structural measurements are anchored to a mature-state chromophore reference rather than to arbitrary sequence positions. Second, the chromophore is decomposed into phenolate, bridge and imidazolinone regions, so the model can distinguish electrostatic stabilisation of the phenolate from bridge rigidity or imidazolinone packing. Third, residue properties are propagated as channel--signal--region tuples: the activated channel defines the physical contact path, the seed signal defines the physicochemical quantity being transmitted, and the target region defines where that signal acts on the chromophore. Fourth, direct residue-identity shortcuts are removed before model fitting, forcing the predictor to rely on transferable local-environment descriptors rather than memorising amino-acid labels. This design makes the selected features mechanistic by construction: a feature such as a charge-related signal reaching the phenolate region through a steric-lock path has a direct physical interpretation before any post-hoc explanation is applied.

\subsection{Mathematical formulation}
For protein $i$, the predictor is a composition of graph construction, state transformation, propagation and regression:
\begin{equation}
\mathcal{A}:(\mathrm{PDB}_i,\mathrm{em}_i)\mapsto \hat{y}_i,
\end{equation}
where $\mathrm{PDB}_i$ is the input structure, $\mathrm{em}_i$ is the emission maximum and $\hat{y}_i$ is predicted QY.

\paragraph{Typed 3D mechanism graph.}
The structure is converted into a typed residue graph
\begin{equation}
G_i=(V_i,\{E_i^{(\tau)}\}_{\tau\in\mathcal{T}_{\mathrm{cand}}},X_i,W_i),
\end{equation}
where $V_i=\{v_1,\ldots,v_{n_i}\}$ is the residue-node set, $X_i\in\mathbb{R}^{n_i\times p}$ stores physicochemical attributes, $E_i^{(\tau)}$ is the edge set for candidate interaction type $\tau$ and $W_i^{(\tau)}$ stores edge weights. We distinguish candidate physical edge annotations from the propagation channels used for prediction. The candidate annotation set is
\begin{equation}
\begin{aligned}
\mathcal{T}_{\mathrm{cand}}=\{&\mathrm{spatial},\mathrm{steric},\mathrm{hydrophobic},\\
&\mathrm{hbond},\mathrm{electrostatic},\mathrm{aromatic}\}.
\end{aligned}
\end{equation}
For the OpenFold3-predicted structures used in this benchmark, the activated propagation channels are
\begin{equation}
\mathcal{T}_{\mathrm{on}}=\{\mathrm{steric},\mathrm{hydrophobic}\}.
\end{equation}
Thus, an annotated edge records a possible physical relation between residues, whereas an activated channel defines the graph path actually used for signal propagation.

\paragraph{Mature-state chromophore registration.}
The immature precursor is represented as
\begin{equation}
 c_i^{(0)}=\mathrm{triad}(X\!-\!\mathrm{Tyr}\!-\!\mathrm{Gly}),
\end{equation}
and the mature chromophore state is obtained by a rule-based transition
\begin{equation}
 c_i^{(t+1)}=\mathcal{R}\!\left(c_i^{(t)},G_i\right),\qquad c_i^\ast=c_i^{(T)}=\cro_i.
\end{equation}
Here $\mathcal{R}$ denotes the implemented cyclization, oxidation and dehydration registration rules. In this work, the purpose of $\mathcal{R}$ is to define a consistent mature CRO anchor for graph features.

\paragraph{CRO regional decomposition.}
The mature chromophore is decomposed into disjoint functional regions,
\begin{equation}
\cro_i=C_i^{\mathrm{phenolate}}\;\dot{\cup}\;C_i^{\mathrm{bridge}}\;\dot{\cup}\;C_i^{\mathrm{imidazolinone}},
\end{equation}
with $\mathcal{R}_{\cro}=\{\mathrm{phenolate},\mathrm{bridge},\mathrm{imidazolinone}\}$. This partition lets the model distinguish phenolate electrostatic stabilisation from bridge rigidity or imidazolinone packing.

\paragraph{Edge-specific signal propagation.}
Each non-chromophore residue $u$ emits a 19-dimensional seed vector

\begin{equation}
 s_u=(s_{u1},s_{u2},\ldots,s_{u,19})\in\mathbb{R}^{19},
\end{equation}
covering steric bulk, flexibility risk, rotatable-bond burden, charge, hydrogen-bond donor/acceptor capacity, aromaticity and hydrophobicity. In the present OpenFold3-based implementation, signal propagation is restricted to the two structurally robust channels $\mathcal{T}_{\mathrm{on}}=\{\mathrm{steric},\mathrm{hydrophobic}\}$. Hydrogen-bond, electrostatic and proton-transfer propagation channels are not activated because the input OpenFold3-predicted PDB structures do not provide hydrogen atoms, ordered water networks or sufficiently reliable local electrostatic field geometry. These channels are therefore treated as reserved physical channels that can be reactivated for X-ray structures or MD ensembles. For activated channel $\tau\in\mathcal{T}_{\mathrm{on}}$ and CRO region $r$, residue $u$ receives the graph-decay weight

\begin{equation}
K_{\tau}(u,r)=\exp[-\lambda_{\tau}d_{\tau}(u,r)]\,\psi_{\tau}(u,r),
\end{equation}
where $d_{\tau}(u,r)$ is the typed-graph distance from $u$ to region $r$, $\lambda_{\tau}$ is the channel-specific decay term and $\psi_{\tau}(u,r)$ represents edge-compatibility along the mechanism path. The enrichment of seed $a$ through channel $\tau$ into region $r$ is
\begin{equation}
F_{a,\tau,r}(G_i)=
\frac{\sum_{u\in V_i\setminus \cro_i}s_{ua}K_{\tau}(u,r)}
{\sum_{u\in V_i\setminus \cro_i}K_{\tau}(u,r)+\varepsilon}.
\label{eq:enrichment}
\end{equation}
With 19 seed signals, two activated propagation channels and three CRO regions, $19\times2\times3=114$ enrichment features are produced. Seven chromophore-clamp descriptors are appended,
\begin{equation}
\begin{aligned}
\phi_i
=&\Big[
\{F_{a,\tau,r}(G_i): a=1,\ldots,19,\\
&\tau\in\mathcal{T}_{\mathrm{on}},
\ r\in\mathcal{R}_{\cro}\},
\kappa_i^{(1)},\ldots,\kappa_i^{(7)}
\Big]\in\mathbb{R}^{121}.
\end{aligned}
\end{equation}

\paragraph{Non-identity feature pool.}
The final mechanism feature pool is
\begin{equation}
\mathcal{F}_{\mathrm{nonID}}=
\left[
\left(\bigcup_{m\in\mathcal{M}}\mathcal{F}_m\right)
\cup\mathcal{F}_{\mathrm{clamp}}
\right]
\setminus\mathcal{F}_{\mathrm{identity}},
\end{equation}
where $\mathcal{M}$ includes steric, hydrophobic, charge-related,
hydrogen-bond-capable, aromatic and solvent-related seed-signal families.
The set $\mathcal{F}_{\mathrm{identity}}$ removes direct residue-identity
shortcuts, including \texttt{is\_Tyr}, \texttt{is\_Phe} and
\texttt{is\_His}. The implementation retains
$|\mathcal{F}_{\mathrm{nonID}}|=52$ features; the 121-to-52
filtering logic is detailed in Table~\ref{tab:appendix_feature_pool}.

\paragraph{Band-specific selection and prediction.}
Band assignment is deterministic for proteins within the predefined modelling bands:
\begin{equation}
b_i=\begin{cases}
\mathrm{GFP\text{-}like}, & 500\leq \mathrm{em}_i<560,\\
\mathrm{Red}, & 580\leq \mathrm{em}_i<610,\\
\mathrm{Far\text{-}red}, & \mathrm{em}_i\geq610.
\end{cases}
\end{equation}
Proteins with emission maxima outside these intervals were excluded from the band-specific benchmark before model fitting, so no sample was routed through an undefined band. Within each training fold and band $b$, features are ranked by
\begin{equation}
\rho_{j,b}=\left|\mathrm{corr}\left(\phi_{\cdot j},y\right)\right|,
\end{equation}
and the selected set is
\begin{equation}
S_b=\operatorname{TopK}_{j\in\mathcal{F}_{\mathrm{nonID}}}(\rho_{j,b}),\qquad K=25.
\end{equation}
The prediction is routed through the corresponding band model:
\begin{equation}
\hat{y}_i=f_{b_i}\!\left(\phi_{i,S_{b_i}}\right),
\end{equation}
where $f_b$ is the ExtraTrees regressor for band $b$ \cite{Geurts2006,Pedregosa2011}.

\paragraph{Top-K screening.}
Training-fold quantiles define bright and dark labels,
\begin{align}
q_{90}&=Q_{0.90}(y_{\mathrm{train}}), &
q_{10}&=Q_{0.10}(y_{\mathrm{train}}),\\
B_i&=\mathbb{I}[y_i\ge q_{90}], &
D_i&=\mathbb{I}[y_i\le q_{10}].
\end{align}
For the $K$ largest and smallest predictions, denoted $\operatorname{TopK}_{+}(\hat{y})$ and $\operatorname{TopK}_{-}(\hat{y})$, the screening metrics are
\begin{align}
\mathrm{Bright\;P@K}&=\frac{1}{K}\sum_{i\in\operatorname{TopK}_{+}(\hat{y})}B_i,\\
\mathrm{Dark\;P@K}&=\frac{1}{K}\sum_{i\in\operatorname{TopK}_{-}(\hat{y})}D_i.
\end{align}

\begin{table}[t]
\centering
\caption{Core algorithmic modules and their implementation in the proposed method.}
\label{tab:algorithm}
\small
\begin{tabular}{p{0.30\columnwidth}p{0.60\columnwidth}}
\toprule
Module & Implementation detail\\
\midrule
Typed 3D graph & Residue nodes with candidate spatial, steric, hydrophobic, hydrogen-bond, electrostatic and aromatic annotations.\\
Mature CRO registration & Rule-based transition from X--Tyr--Gly precursor to a mature-state CRO anchor for downstream geometry.\\
CRO partition & Separate phenolate, bridge and imidazolinone target regions.\\
Signal propagation & Channel--signal--region propagation: 19 residue signals are transmitted through two OpenFold3-robust activated channels, steric and hydrophobic, into three CRO regions.\\
Non-identity pool & 52 edge/clamp-geometry features after removing direct amino-acid identity shortcuts.\\
Band model & Training-fold-only feature selection and per-band ExtraTrees regression.\\
\bottomrule
\end{tabular}
\end{table}

\subsection{Edge-family construction and implementation details}
The feature construction is not a generic graph featurisation step. It is organised by physically distinct propagation channels and seed-signal families. In the present OpenFold3-based setting, steric and hydrophobic contact channels are activated because their geometric definitions are robust to the absence of hydrogens and explicit solvent. Within these channels, the seed signals still encode bulky contacts, flexibility risk, rotatable-bond burden, charge-related side-chain properties, hydrogen-bond donor/acceptor capacity, aromaticity and hydrophobicity. Direct residue-identity strings are removed after candidate construction, so aromatic or hydrophobic features are retained only when they represent physicochemical signal classes rather than amino-acid labels.

This channel--signal design gives the algorithm a useful inductive bias. A charge-related seed propagated through a steric-lock path to the phenolate region is not treated as equivalent to a rotatable-bond burden propagated through a hydrophobic-contact path to the bridge region, even if both residues are close in Euclidean distance. Instead, the feature name encodes a mechanism tuple,
\begin{equation}
\text{feature}=(\text{channel},\text{seed signal},\text{CRO region}),
\end{equation}
which is why the selected features can later be interpreted as concrete mechanistic motifs.

\subsection{Implementation-level propagation computation}
Eq.~\ref{eq:enrichment} defines the conceptual channel--signal--region enrichment form. In the implementation, this quantity is realised by a finite-step, chromophore-local message-passing readout, followed by a region-specific weighted readout at the phenolate, bridge or imidazolinone CRO region. Residue--residue distances are computed from atom coordinates rather than only from residue centroids:
\begin{equation}
d(u,v)=\min_{a\in A_u,\,b\in A_v}\left\|\mathbf{r}_{u,a}-\mathbf{r}_{v,b}\right\|_2,
\end{equation}
where $A_u$ and $A_v$ are atom sets for residues $u$ and $v$. The residue centre used for local filtering is
\begin{equation}
\mathbf{c}_u=\frac{1}{|A_u|}\sum_{a\in A_u}\mathbf{r}_{u,a}.
\end{equation}
For the activated channels, steric-lock contacts are weighted by
\begin{equation}
w^{(\mathrm{steric})}_{uv}=\exp[-d(u,v)/3.0],
\end{equation}
and hydrophobic contacts by
\begin{equation}
w^{(\mathrm{hydrophobic})}_{uv}=\frac{1}{1+d(u,v)}.
\end{equation}
Only nodes within a chromophore-local neighbourhood are propagated:
\begin{equation}
V_i^{\mathrm{local}}=\{u\in V_i\setminus \cro_i:\|\mathbf{c}_u-\mathbf{c}_{\cro}\|_2<12.0\;\text{\AA}\}.
\end{equation}
For activated channel $\tau$, the node state is initialised as $\mathbf{h}^{(0)}_{u,\tau}=\mathbf{s}_u$ and updated for $T=2$ propagation steps:
\begin{equation}
\mathbf{h}^{(t+1)}_{u,\tau}=\frac{\mathbf{h}^{(t)}_{u,\tau}+\sum_{v\in\mathcal{N}_{\tau}(u)} w^{(\tau)}_{uv}\alpha_u\mathbf{h}^{(t)}_{v,\tau}}{1+0.1\sum_{v\in\mathcal{N}_{\tau}(u)} w^{(\tau)}_{uv}\alpha_u},
\end{equation}
where
\begin{equation}
\alpha_u=\exp\left[-\|\mathbf{c}_u-\mathbf{c}_{\cro}\|_2/5.0\right].
\end{equation}
The final readout for CRO region $r$ uses a region-specific distance weight,
\begin{equation}
\beta_{u,r}=\exp\left[-\|\mathbf{c}_u-\mathbf{c}_{r}\|_2/3.0\right],
\end{equation}
and computes
\begin{equation}
F_{a,\tau,r}=\frac{\sum_{u:\,\beta_{u,r}>0.05}\beta_{u,r}h^{(T)}_{u,\tau,a}}{\sum_{u:\,\beta_{u,r}>0.05}\beta_{u,r}+\varepsilon}.
\end{equation}
Thus, the implementation first propagates residue physicochemical signals over activated local contact graphs and then reads out the propagated state at phenolate, bridge and imidazolinone CRO regions.

\subsection{Leakage control and distributional diagnostics}
The random-CV experiment uses out-of-fold prediction throughout. For seed \(s\), the QY values are first converted into quantile bins,
\begin{equation}
z_i=\operatorname{qcut}(y_i,Q),\qquad Q=\min(5,\lfloor n/5\rfloor),
\end{equation}
and stratified folds are built from \(z_i\). Feature selection, band-specific model fitting and bright/dark thresholds are recomputed inside each training fold only. Thus, the held-out fold does not influence the selected feature set, the fitted regressor or the Top-K threshold.

We additionally monitor prediction compression,
\begin{equation}
C=\frac{\operatorname{sd}(\hat{y})}{\operatorname{sd}(y)},
\end{equation}
where values much smaller than one indicate collapse toward the mean. This diagnostic is useful for QY prediction because a model that only predicts mid-range QY may have tolerable MAE but poor value for bright/dark candidate discovery.

\section{Results}
The evaluation was designed to test whether chromophore-centred environment features improve both prediction and mechanism-level interpretation. We therefore report random-CV regression, homology-controlled generalisation, bright/dark Top-K screening and the stability of selected channel--signal--region features.

\subsection{Random-CV QY prediction}
Under random cross-validation, the mechanism graph model achieved the strongest primary model-based performance (Table~\ref{tab:random}; Fig.~\ref{fig:regimes}). Its mean Pearson correlation was $R=0.772\pm0.008$, compared with $R=0.734$ for ESM-C, $R=0.731$ for SaProt and $R=0.632$ for Band mean. It also had the lowest MAE ($0.131\pm0.002$), indicating that chromophore-centred mechanism features improve both ranking and absolute regression error.

The same pattern appeared in candidate screening. The proposed method ranked first in the most stringent bright-screening setting, with Bright P@5 of 0.704 versus 0.680 for ESM-C, 0.640 for SaProt and 0.160 for Band mean. For Dark P@5, it reached 0.536, exceeding ESM-C (0.507), SaProt (0.453) and Band mean (0.227). At larger K, PLM baselines remained competitive on bright retrieval, but the mechanism graph model stayed in the strongest tier while retaining interpretability.

\begin{table}[t]
\centering
\caption{Random-CV regression performance. Means for the proposed method are reported across five seeds; baseline values are averaged across random-seed runs.}
\label{tab:random}
\begin{tabular}{lcc}
\hline
Method & Random $R$ & Random MAE\\
\hline
Band mean & 0.632 $\pm$ 0.002 & 0.167 $\pm$ 0.000\\
ESM-C & 0.734 $\pm$ 0.005 & 0.143 $\pm$ 0.001\\
SaProt & 0.731 $\pm$ 0.002 & 0.146 $\pm$ 0.000\\
\textbf{Mechanism graph} & 0.772 $\pm$ 0.008 & 0.131 $\pm$ 0.002\\
\hline
\end{tabular}

\end{table}

\subsection{Homology-controlled generalisation}
\begin{figure}[t]
    \centering
    \includegraphics[width=\columnwidth]{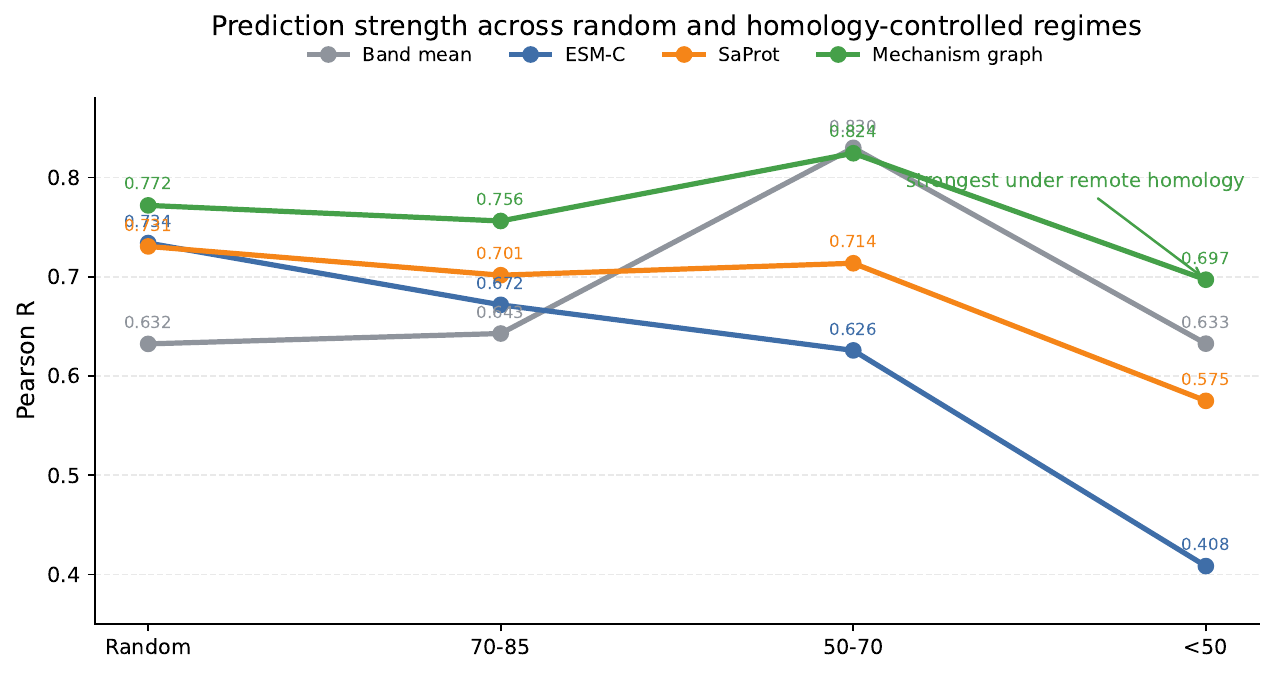}
    \caption{Prediction strength across random and fixed 5-mer-Jaccard homology-controlled regimes. The x-axis bucket labels 70--85, 50--70 and $<50$ are compact labels for $J_5$ intervals 0.70--0.85, 0.50--0.70 and $<0.50$, respectively, where $J_5$ is the maximum 5-mer Jaccard similarity from each held-out protein to the fixed training set. These labels do not denote global sequence-identity percentages.}
    \label{fig:regimes}
\end{figure}

The fixed 5-mer-Jaccard homology-controlled evaluation tested whether the models could transfer from a training set composed of proteins with close internal neighbours to held-out proteins with progressively weaker similarity to that training set (Table~\ref{tab:homology}; Fig.~\ref{fig:regimes}). Bucket membership was determined by each test protein's maximum 5-mer Jaccard similarity to the fixed training set, not by random fold assignment.

In the 0.70--0.85 bucket, the method reached $R=0.756$, outperforming Band mean (0.643), ESM-C (0.672) and SaProt (0.701). In the 0.50--0.70 bucket, it reached $R=0.824$, essentially matching Band mean (0.830) while clearly exceeding ESM-C (0.626) and SaProt (0.714). In the most remote bucket ($<0.50$), it was clearly strongest with $R=0.697$, compared with 0.633 for Band mean, 0.575 for SaProt and 0.408 for ESM-C.

This pattern is important because the $<0.50$ bucket is the closest evaluation setting to novel FP design: held-out proteins share limited 5-mer motif content with the fixed training set, even though they still belong to the broader fluorescent-protein structural family. In this regime, generic representations degraded more sharply, whereas chromophore-region mechanism features remained comparatively transferable.

\begin{table}[t]
\centering
\caption{Pearson correlation under fixed 5-mer-Jaccard homology-controlled evaluation. For compactness and consistency with Fig.~\ref{fig:regimes}, bucket labels are shown as 70--85, 50--70 and $<50$, corresponding to $J_5$ intervals 0.70--0.85, 0.50--0.70 and $<0.50$, respectively. These labels do not denote global sequence-identity percentages.}
\label{tab:homology}
\begin{tabular}{lccc}
\toprule
Method & 70--85 & 50--70 & $<50$\\
\midrule
Band mean & 0.643 & 0.830 & 0.633\\
ESM-C & 0.672 & 0.626 & 0.408\\
SaProt & 0.701 & 0.714 & 0.575\\
Mechanism graph & 0.756 & 0.824 & 0.697\\
\bottomrule
\end{tabular}
\end{table}

\subsection{Remote-homology Top-K screening}
\begin{figure*}[!t]
    \centering
    \includegraphics[
        width=0.90\textwidth,
        height=0.50\textheight,
        keepaspectratio,
        trim=0.20in 0.12in 0.20in 0.12in,
        clip
    ]{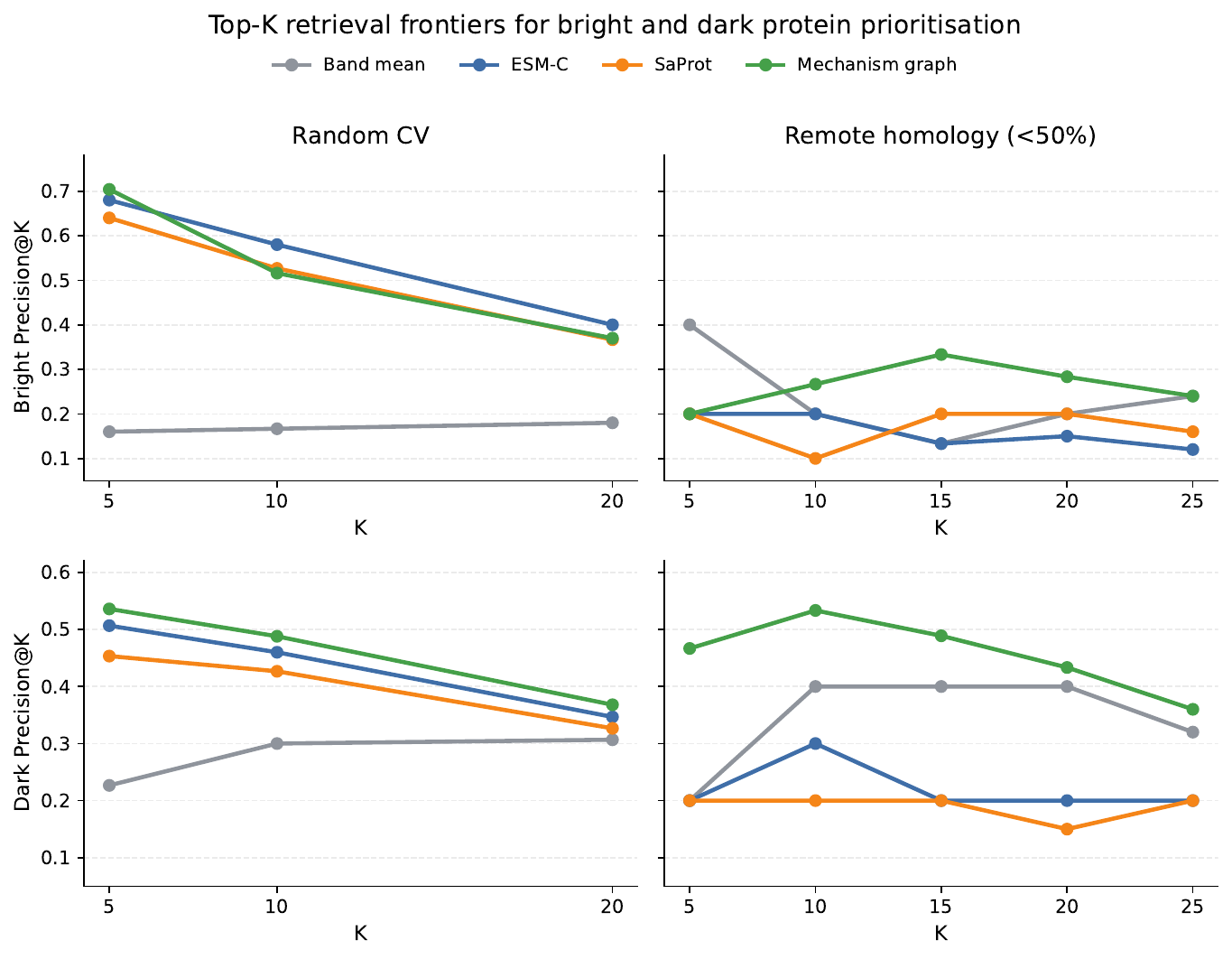}
    \vspace{-0.6em}
    \caption{Top-K retrieval frontiers for bright and dark proteins. Left: random CV. Right: the most remote fixed-split bucket, where each held-out protein has maximum 5-mer Jaccard similarity $m_k<0.50$ to the fixed training set. The embedded panel label ``Remote homology ($<50\%$)'' is a compact label for this $J_5<0.50$ bucket and does not denote pairwise sequence identity.}
    \label{fig:topk}
    \vspace{-0.8em}
\end{figure*}

Candidate prioritisation under remote homology is a practical test for FP engineering because novel designs often lack close labelled neighbours. In the most remote 5-mer-Jaccard bucket ($m_k<0.50$), the proposed method produced the best bright precision at $K=10,15,20$ and 25, and the best dark precision at all tested $K$ values from 5 to 25 (Fig.~\ref{fig:topk}). For example, Dark P@10 reached 0.533, compared with 0.400 for Band mean, 0.300 for ESM-C and 0.200 for SaProt. Bright P@15 reached 0.333, compared with 0.133 for Band mean, 0.133 for ESM-C and 0.200 for SaProt. Thus, the method is not only a stronger remote-homology regressor but also a useful screening tool for fixed-split extrapolation.

\subsection{Stable selected features expose band-specific mechanism signatures}
\begin{figure*}[!t]
    \centering
    \includegraphics[
        width=0.94\textwidth,
        height=0.48\textheight,
        keepaspectratio
    ]{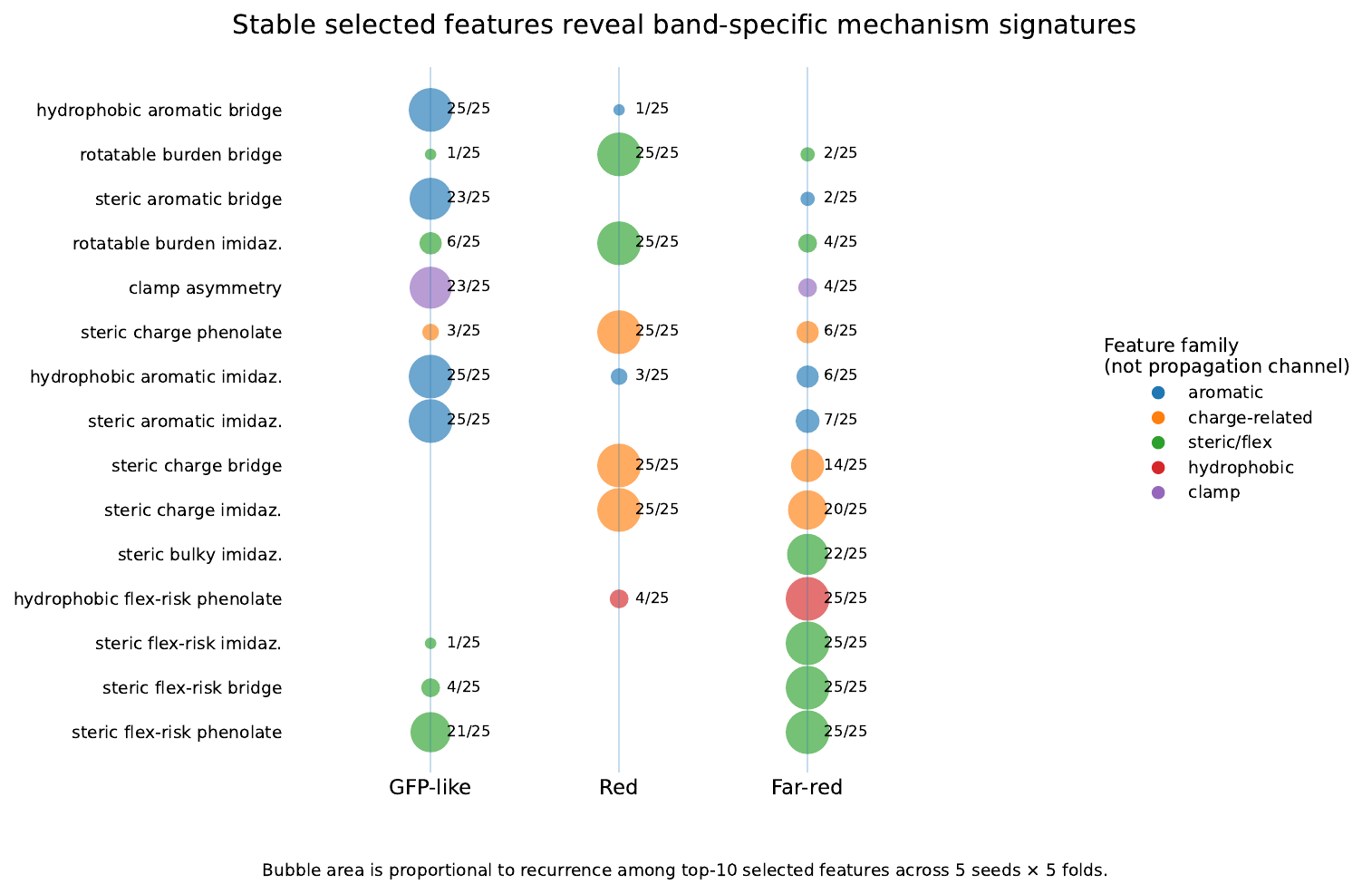}
    \vspace{-0.5em}
    \caption{Stable selected mechanism descriptors across seeds and folds. The three columns correspond to the GFP-like, Red and Far-red band-specific models. Each bubble denotes a channel--signal--region descriptor or a local clamp descriptor that repeatedly appeared among the top-10 selected features across five seeds and five folds; bubble area is proportional to recurrence. Colours denote feature families, not propagation channels. The activated propagation channels are steric and hydrophobic, whereas feature families describe the physicochemical seed signal or clamp descriptor carried through those channels. The label \emph{clamp asymmetry} denotes the local chromophore-clamp asymmetry feature after removing implementation-specific residue numbering.}
    \label{fig:mechanisms}
    \vspace{-0.6em}
\end{figure*}

The feature-stability map is central to the interpretability claim and is therefore retained as a main-text figure (Fig.~\ref{fig:mechanisms}). Each point corresponds to a concrete selected feature, not to a post-hoc attribution score. Bubble size indicates how often that feature appeared among the top-10 selected features across five seeds and five folds.

The feature-stability map should be interpreted as a band-specific summary of repeatedly selected mechanism descriptors, not as a per-protein attribution map. Each bubble denotes one channel--signal--region descriptor, or one local clamp descriptor, that repeatedly entered the top-ranked feature set during training-fold feature selection. Thus, the plot asks which physicochemical signals are consistently useful for QY prediction within each emission class.

The resulting patterns are biologically coherent and consistent with known fluorescent-protein photophysics. In GFP-like proteins, hydrophobic/aromatic and steric/aromatic signals into the bridge and imidazolinone readouts were repeatedly selected, together with clamp asymmetry. This agrees with experimental and chromophore-model studies showing that GFP-like fluorescence is enhanced when chromophore torsional motion is restricted and the emitting state is rigidified by the protein environment \cite{Follenius2003,Ferreira2022,Park2016}. In red proteins, charge-related signals and rotatable-burden descriptors were repeatedly selected around the phenolate, bridge and imidazolinone readouts. This is consistent with studies of red fluorescent proteins in which local electric fields, electrostatic/hydrogen-bond interactions and chromophore planarity tune fluorescence quantum yield \cite{Drobizhev2021,Bindels2017}. In far-red proteins, flex-risk and bulky-contact descriptors were most stable, especially around bridge and imidazolinone readouts. This matches structural and photophysical work showing that far-red fluorescence is strongly affected by chromophore isomerization, planarity and steric restriction of torsional relaxation \cite{Pletnev2008,Petersen2003,Legault2022}. These agreements do not prove that each selected descriptor is individually causal, but they show that the model-selected signatures recover known physical themes in a band-specific manner.

\FloatBarrier

\subsection{Audited ablation analysis}
We performed an audited ablation analysis to test whether the prediction came from the full chromophore-centred representation or from a single shortcut component. The audited experiment used 54 conditions, five seeds and five folds per seed, giving 143{,}370 out-of-fold predictions. Label shuffling collapsed performance to $R=0.006$, confirming that the model captured non-random structure--QY signal (Table~\ref{tab:ablation_audited}).

The ablation results support the full 52-feature representation as a robust pre-specified mechanism model. The propagated enrichment features were the dominant predictive component: enrichment-only features reached $R=0.769$, whereas the seven V54 clamp descriptors alone reached $R=0.692$. Adding the clamp descriptors to enrichment features gave the full model ($R=0.772$), indicating that V54 descriptors provide standalone geometric information and a small positive marginal trend, while the main predictive signal is carried by propagated chromophore-centred enrichment. Single-channel controls were weaker than the full representation after removing clamp descriptors, supporting the use of a combined physicochemical enrichment representation rather than a single isolated channel.

Region controls further showed that the pre-specified region-resolved construction can be interpreted as a chromophore-centred readout frame. Sample-wise scrambling of regional readouts reduced performance ($R=0.722$), indicating that consistent chromophore-centred readout semantics are important for the propagated representation. The separate retrained remote-ablation subset was treated only as a qualitative sensitivity check because of its small test size, and it was not used to replace the primary homology-controlled benchmark reported above.

\begin{table*}[!t]
\centering
\small
\caption{Audited ablation results used to support the full pre-specified mechanism model. The table reports representative conditions from the audited 54-condition study. ``Features'' denotes the candidate feature count before training-fold Top-25 selection, except for controls using no mechanism features.}
\label{tab:ablation_audited}
\small
\begin{tabular}{p{0.39\textwidth}r r p{0.36\textwidth}}
\toprule
Condition & Features & $R$ & Interpretation\\
\midrule
Shuffle QY labels & 52 & 0.006 & Null control; performance collapses when labels are randomised.\\
Band mean & 0 & 0.627 & Emission-band baseline.\\
Emission maximum only & 1 & 0.639 & Continuous emission metadata alone.\\
Global 52-feature model & 52 & 0.675 & Mechanism features without band-specific routing.\\
V54 clamp only & 7 & 0.692 & Local chromophore-clamp geometry alone.\\
Steric-channel only, no clamp & 21 & 0.756 & A single activated channel is insufficient.\\
Hydrophobic-channel only, no clamp & 24 & 0.767 & The second activated channel is informative but incomplete alone.\\
Enrichment only, no clamp & 45 & 0.769 & Propagated enrichment carries the dominant predictive signal.\\
Full pre-specified mechanism model & 52 & 0.772 & Region-resolved enrichment plus V54 clamp descriptors.\\
\bottomrule
\end{tabular}

\vspace{-0.6em}
\end{table*}
\FloatBarrier

\section{Discussion}
\noindent
The proposed algorithm was designed around the idea that QY should be predicted as a local chromophore-environment problem rather than as ordinary sequence-to-property regression. The results support this premise. The method improves random-CV regression over Band mean, ESM-C and SaProt, and its strongest advantage appears under remote-homology evaluation, where labelled neighbours and generic embeddings are less reliable.

The audited ablation analysis refines this interpretation. The useful signal is not attributable to a single hand-crafted descriptor, seed family or propagation channel. Instead, performance is mainly supported by propagated chromophore-centred enrichment features combined with emission-aware, band-specific learning. The region-resolved representation is retained as the pre-specified mechanism construction because it provides concrete channel--signal--region features, while the audited ablation results support the use of the full chromophore-centred representation rather than a single shortcut component.

Several algorithmic details are essential and should not be reduced to ordinary feature engineering. First, the typed 3D graph changes the prediction problem from residue-identity lookup to propagation over physical interaction channels. Second, mature-state chromophore registration gives a consistent chemical anchor for all downstream geometric features. Third, CRO regional decomposition provides a region-resolved readout frame for phenolate, bridge and imidazolinone signals, allowing selected features to remain interpretable as concrete channel--signal--region descriptors. Fourth, edge-specific propagation separates the activated physical contact channel from the physicochemical seed signal, so that steric-lock and hydrophobic-contact paths can carry distinct charge-related, hydrogen-bond-capable, aromatic, flexibility-risk and bulky-contact information. Fifth, the non-identity feature pool removes direct amino-acid shortcuts, making the learned inputs closer to transferable physical descriptors than to sequence-family tags.

The restored feature-stability map is especially important because it connects the algorithm to interpretable photophysical hypotheses already suggested by structural and spectroscopic studies of fluorescent proteins. The GFP-like signatures emphasise aromatic packing and clamp-related rigidity, matching the established view that restricted chromophore torsion and a rigid local pocket favour radiative decay. The red-protein signatures emphasise charge-related and rotatable-burden descriptors, consistent with local electric-field and chromophore-planarity mechanisms reported for red fluorescent proteins. The far-red signatures emphasise flex-risk and bulky-contact descriptors, consistent with the role of chromophore isomerization, planarity and steric restriction in far-red emission. These patterns are not post-hoc feature attributions; they are the feature families repeatedly selected inside training folds. They should therefore be interpreted as stable, literature-consistent mechanism signatures rather than as proof that any single descriptor is independently causal. Accordingly, we treat the region-resolved descriptors as mechanistic readout coordinates that organise chromophore-centred signals, while avoiding over-interpretation of any single CRO region label.

The maturation component should be interpreted carefully. It is valuable because it gives a consistent mature-CRO anchor for feature extraction, but the current implementation is rule-based and lightweight rather than a detailed chemical kinetics or quantum-chemical simulation. Future work could replace or augment this step with more explicit chemical-state modelling, learned propagation kernels, uncertainty-aware structural ensembles, or reactivated hydrogen-bond/electrostatic/proton-transfer channels when experimental structures or MD trajectories provide hydrogen atoms, ordered water networks and reliable local electrostatic geometry. This conservative description is important: the novelty here is not a complete reaction simulator, but a graph-learning framework that uses a mature chromophore reference state to define region-specific signal propagation.

For FP engineering, the main value of the method is the combination of predictive performance, remote-homology transfer and interpretable mechanism signatures. It can prioritise bright candidates, identify likely dark proteins and report which chromophore-region interactions drive the prediction. That combination makes it complementary to generic PLM baselines and well suited to design settings in which mechanistic guidance matters.

\FloatBarrier

\section*{Funding}
This research received no specific grant from any funding agency in the public, commercial or not-for-profit sectors.

\section*{Acknowledgements}
The authors thank the maintainers of FPbase and the Protein Data Bank for making fluorescent-protein metadata and structural resources publicly available, and the developers of OpenFold3, ESM-C, SaProt and scikit-learn for releasing the tools used in this study.

\clearpage
\onecolumn
\appendix

\section{Implementation details of channel--signal--region features}
\subsection{Activated channels and reserved channels}
The implementation separates candidate physical annotations from activated propagation channels. Candidate annotations can include spatial proximity, steric contact, hydrophobic contact, hydrogen-bond-related geometry, electrostatic relation and aromatic contact. However, the OpenFold3-predicted PDB structures used in this benchmark do not provide hydrogen atoms, explicit water molecules or reliable local electrostatic field geometry. Therefore, only the two geometry-robust contact channels are activated for propagation:
\begin{equation}
\mathcal{T}_{\mathrm{on}}=\{\mathrm{steric},\mathrm{hydrophobic}\}.
\end{equation}
Hydrogen-bond, electrostatic and proton-transfer channels are treated as reserved physical channels. They are not discarded conceptually; rather, they are not activated in the present benchmark because their reliable use would require experimental structures with hydrogens, explicit solvent, protonation-state assignment or MD ensembles.

\subsection{Seed-signal families}
Each residue emits physicochemical seed signals, and these signals are grouped into six interpretable families before the non-identity feature pool is constructed:
\begin{equation}
\mathcal{M}=
\{\mathrm{steric},\mathrm{hydrophobic},\mathrm{charge},
\mathrm{hbond},\mathrm{aromatic},\mathrm{solvent}\}.
\end{equation}
The important distinction is that these are seed-signal families, not necessarily independent propagation channels. For example, a charge-related seed can be propagated through a steric-lock path or through a hydrophobic-contact path, producing different channel--signal--region features. This is why the representation can retain charge-related, hydrogen-bond-capable and aromatic information even though only steric and hydrophobic contact channels are activated in the OpenFold3-based setting.

\subsection{Feature counting and non-identity filtering}
The full chromophore-centred representation contains 121 candidate
features. These consist of 57 steric-channel enrichment features,
57 hydrophobic-channel enrichment features and seven local
chromophore-clamp descriptors:
\begin{equation}
(19\times3)_{\mathrm{steric}}
+
(19\times3)_{\mathrm{hydrophobic}}
+
7_{\mathrm{clamp}}
=121.
\end{equation}
Here the factor 19 denotes residue seed signals and the factor
3 denotes the phenolate, bridge and imidazolinone CRO regions.

The reduction from 121 to 52 features is performed in two steps.
First, the implementation keeps only seed signals assigned to six
physicochemical signal families plus the local chromophore-clamp descriptors.
This family mapping covers 73 columns:
\begin{equation}
66_{\mathrm{family\ enrichment}}+7_{\mathrm{clamp}}=73.
\end{equation}
The remaining 48 enrichment columns are not used because they
are either direct residue-identity signals outside the retained
families or auxiliary signals not included in the final mechanism
pool. Second, direct amino-acid identity shortcuts within the
73-column family pool are removed. These include aromatic-family
\texttt{is\_PHE}, \texttt{is\_TYR} and \texttt{is\_HIS} features in both
activated channels, and solvent-family \texttt{is\_ALA} features in
the hydrophobic channel:
\begin{equation}
73-21=52.
\end{equation}

The final non-identity pool therefore contains 45 enrichment
features and seven clamp descriptors:
\begin{equation}
52 = 45_{\mathrm{nonID\ enrichment}}+7_{\mathrm{clamp}}.
\end{equation}

\begin{table}[H]
\centering
\caption{Composition of the final 52-feature non-identity pool. Each
number in parentheses counts the three CRO regions unless otherwise
noted. Clamp descriptors are channel-independent local geometry
features, not propagated steric-channel features.}
\label{tab:appendix_feature_pool}
\small
\begin{tabular}{p{0.15\textwidth}p{0.26\textwidth}p{0.27\textwidth}p{0.20\textwidth}r}
\toprule
Family & Steric channel & Hydrophobic channel & Channel-independent & Total\\
\midrule
Steric &
bulky, flex-risk, rotatable-bond burden (9) &
-- &
-- & 9\\
Hydrophobic &
-- &
hydrophobic, bulky contact (6) &
-- & 6\\
Charge-related &
charge (3) &
charge (3) &
-- & 6\\
H-bond-capable &
donor, acceptor (6) &
donor, acceptor (6) &
-- & 12\\
Aromatic &
aromatic signal (3) &
aromatic signal (3) &
-- & 6\\
Solvent-related &
-- &
flex-risk, total contact burden (6) &
-- & 6\\
Clamp &
-- &
-- &
local bridge/phenolate/imidazolinone/asymmetry descriptors (7) & 7\\
\midrule
Total & & & & 52\\
\bottomrule
\end{tabular}
\end{table}
\FloatBarrier

\FloatBarrier
\subsection{Intrinsic interpretation}
The model interpretation is intrinsic to the representation. Each selected feature is already a mechanism tuple,
\begin{equation}
\mathrm{feature}=(\mathrm{channel},\mathrm{seed\ signal},\mathrm{CRO\ region}),
\end{equation}
rather than an attribution score assigned after model fitting. Thus, stable selected features directly identify which physical signal reaches which chromophore region through which activated contact path. This is the basis for interpreting GFP-like models as aromatic-packing dominated, Red models as charge/rotatable-burden dominated and Far-red models as flex-risk/bulky-contact dominated.

\subsection{Auxiliary V54 stress test}
Because the structural inputs include OpenFold3-predicted models, we also ran a small stress test to ask whether V54 clamp descriptors buffer simulated errors in propagated enrichment features. The training folds were kept clean, whereas enrichment features in the held-out fold were perturbed by Gaussian noise, feature dropout or a bad-structure subset perturbation. V54 features were not perturbed. The comparison was restricted to enrichment-only, full and V54-only feature sets, so the test was used as a conservative robustness check rather than as a primary model-selection experiment.

The full model showed a directionally consistent but statistically inconclusive buffer effect relative to enrichment-only features (Table~\ref{tab:v54_stress}). Across Gaussian-noise and dropout settings, Buffer$_R$ was small ($+0.002$ to $+0.005$) and paired bootstrap confidence intervals included zero. Therefore, V54 clamp descriptors are best interpreted as auxiliary local-geometry descriptors with standalone predictive information, not as a demonstrated correction mechanism for structure-prediction errors.

\begin{table}[H]
\centering
\caption{V54 structure-error stress test. Buffer$_R$ is the difference between enrichment-only degradation and full-model degradation under the same simulated enrichment-feature corruption.}
\label{tab:v54_stress}
\small
\begin{tabular}{lccc}
\toprule
Stress condition & Enrichment-only $R$ & Full $R$ & Buffer$_R$\\
\midrule
Clean input & 0.7692 & 0.7719 & --\\
Gaussian noise, $\sigma=0.10$ & 0.7469 & 0.7532 & +0.0036\\
Gaussian noise, $\sigma=0.20$ & 0.7350 & 0.7398 & +0.0023\\
Gaussian noise, $\sigma=0.30$ & 0.7238 & 0.7301 & +0.0036\\
Feature dropout, $p=0.10$ & 0.7583 & 0.7630 & +0.0020\\
Feature dropout, $p=0.20$ & 0.7420 & 0.7500 & +0.0052\\
Bad-structure subset, 20\% & 0.7585 & 0.7615 & +0.0004\\
\bottomrule
\end{tabular}

\end{table}

\end{document}